%% file: main.tex
\documentclass[sigconf]{acmart}
\usepackage{amsmath}

\usepackage{xcolor}
\usepackage{tikz}
\usepackage{booktabs}

\newcommand{\short}{IBCI}
\newcommand{\dis}{TIE}
\newcommand{\std}{IIM}
\usepackage{algorithm}
\usepackage{algorithmic}
\usepackage{multirow}
\usepackage{breakurl}
\usepackage[capitalise]{cleveref}

 \AtBeginDocument{%
   \providecommand\BibTeX{{%
     \normalfont B\kern-0.5em{\scshape i\kern-0.25em b}\kern-0.8em\TeX}}}

\copyrightyear{2021} 
\acmYear{2021} 
\setcopyright{acmcopyright}\acmConference[CIKM '21]{Proceedings of the 30th ACM International Conference on Information and Knowledge Management}{November 1--5, 2021}{Virtual Event, QLD, Australia}
\acmBooktitle{Proceedings of the 30th ACM International Conference on Information and Knowledge Management (CIKM '21), November 1--5, 2021, Virtual Event, QLD, Australia}
\acmPrice{15.00}
\acmDOI{xxx/xxx}
\acmISBN{978-1-4503-8446-9/21/11}

\begin{document}
%\fancyhead{}

\title{Neuron Campaign for Initialization Guided by Information Bottleneck Theory}

\author{Haitao Mao}
\authornote{These authors contributed equally to the work.}
\authornote{Work performed during the internship at MSRA.}
\email{2018091709027@std.uestc.edu.cn}
\orcid{1234-5678-9012}
\affiliation{%
  \institution{University of Electronic Science and Technology of China}
  \city{Chengdu}
  \country{China}
}

\author{Xu Chen}
\authornotemark[1]
\authornotemark[2]
\email{sylover@pku.edu.cn}
\affiliation{%
  \institution{Peking University}
  \city{Beijing}
  \country{China}
}

\author{Qiang Fu}
\authornotemark[1]
\email{qifu@microsoft.com}
\affiliation{%
  \institution{Microsoft Research Asia}
  \city{Beijing}
  \country{China}
}

\author{Lun Du}
\authornotemark[1]
\authornote{Corresponding Author.}
  \email{lun.du@microsoft.com}
 \affiliation{%
  \institution{Microsoft Research Asia}
  \city{Beijing}
  \country{China}
}

\author{Shi Han}
\email{shihan@microsoft.com}
\affiliation{%
  \institution{Microsoft Research Asia}
  \city{Beijing}
  \country{China}
}

\author{Domei Zhang}
\email{dongmeiz@microsoft.com}
\affiliation{%
  \institution{Microsoft Research Asia}
  \city{Beijing}
  \country{China}
}

\begin{abstract}
  Initialization plays a critical role in the training of deep neural networks (DNN). 
  Existing initialization strategies mainly focus on stabilizing the training process to mitigate gradient vanish/explosion problems. However, these initialization methods are lacking in consideration about how to enhance generalization ability. The Information Bottleneck (IB) theory is a well-known understanding framework to provide an explanation about the generalization of DNN. Guided by the insights provided by IB theory, we design two criteria for better initializing DNN. And we further design a neuron campaign initialization algorithm to efficiently select a good initialization for a neural network on a given dataset. The experiments on MNIST dataset show that our method can lead to a better generalization performance with faster convergence.
\end{abstract}




\maketitle
\input{introduction}

\input{related_work}

\input{model}

\input{experiment}

\input{conclusion}

%\clearpage
\bibliographystyle{ACM-Reference-Format}
\bibliography{ref}

% \appendix
% \input{src/appendix}

\end{document}

%% file: introduction.tex
\section{Introduction}
Deep Neural Network has built huge success in several fields including computer vision \cite{krizhevsky2012imagenet, he2016deep}, 
natural language understanding \cite{devlin2018bert,floridi2020gpt}, speech recognition \cite{dahl2011context,huang2014deep}, graph mining \cite{yang2020domain,chen2020tssrgcn}, and so on \cite{du2021tabularnet,wang2019tag2vec,wang2019tag2gauss}. However, it is still difficult to train these deep models.

One potential difficulty of training these deep models lies in how to properly initialize the model parameters. \cite{hayou2019impact} suggests that an inappropriate initialization can lead to the gradient vanish/explosion during training procedure and poor generalization performance, like the random initialization adding Gaussian noise with zero-mean and standard deviation equaling 0.01. 

To solve the gradient vanish/explosion problems, existing works have proposed various initialization methods. Xavier initialization \cite{glorot2010understanding} normalizes the output variance in the forward path and the gradient variance during backpropagation under the linear case while He-initialization \cite{he2015delving} is developed similarly for networks with rectifier nonlinearities. 
The above methods are built under various assumptions. To relax the constraint and adopt the same output variance in the forward path, \cite{mishkin2015all} introduces training data and normalizes the variance of its output.
These initialization strategies have achieved great success in mitigating gradient vanish/explosion problems by controlling each layer's response within a proper range. 
However, these works take less consideration about how to enhance model generalization ability in the initialization phase.

In fact, appropriate initialization is indeed advantageous for generalization. Pretrained models obtained by heavy training on large-scale data could be regarded as searching for good initialization parameters like BERT \cite{devlin2018bert} in NLP and \cite{he2020momentum} in CV. It shows great benefits on the generalization ability of downstream tasks.

In this paper, we design a new initialization strategy called Information Bottleneck guided neuron Campaign Initialization (IBCI) that can promote neural network generalization without heavy pre-training.
In more detail, we first generate a large candidate neuron set for campaign by traditional initialization strategies like Xavier. Then, the winning neuron set selects candidate neurons under some criteria and eventually is integrated as the initialization of neural networks. Specifically, we take advantage of the greedy algorithm for the optimal subset selection for practical efficiency. The criteria for selecting the neuron subset are designed according to the Information Bottleneck theory \cite{shwartz2017opening}. 

The IB theory explains DNN's generalization by analyzing the balance of input information maintenance (measured by the mutual information between the input feature $X$ and the latent representation $Z$, i.e., $I(X; Z)$) and target-related information enhancement (measured by the mutual information between $Z$ and the target $Y$, i.e., $I(Z; Y)$) from two perspectives. Regarding the training dynamics, $I(X; Z)$ increases in the early training phase and then decreases while $I(Z; Y)$ keeps maintaining an increasing trend, which indicates that the network memorizes the input information in the beginning and then compresses the unimportant information for generalization. From the perspective of the neural network architecture, front layers near the input focus more on the input information maintenance while rear layers close to the output layer pay more attention to the target-related information enhancement.

As initialization could be considered as a very early stage of training, we design our initialization algorithm according to the aforementioned key insights of the IB theory, i.e., front layer and rear layer focusing more on maximizing I(x,z) and I(z,y), respectively. 
Hence models initialized by our algorithm are much better than traditional initialization in convergence and generalization capacity. The code is available at \url{https://github.com/huanhuqueyue/CIKM-IBCI}.

Our primary contributions can be summarized as follows:
\begin{itemize}
    \item We propose a new initialization perspective on improving model generalization by applying the IB theory. To the best of our knowledge, we are the first to introduce the IB Theory into initialization.
    \item We design a neuron campaign initialization algorithm guided by IB theory to select desired neuron subset from a large set of candidates. The algorithm is both effective and efficient without training procedures. 
    \item Comprehensive experiments conducted on MNIST dataset show that \short{} consistently outperforms other initial methods with higher accuracy and faster convergence.
\end{itemize}

%% file: related_work.tex
\section{Related Work}
The proper initialized model leads to a stable convergence procedure and better convergence result. \cite{glorot2010understanding} proposes the normalized initialization scheme based on the number of nodes of input and output, while He initialization \cite{he2015delving} focuses on more commonly used ReLU. \cite{saxe2013exact} theoretically proves the capability of orthogonal initialization. However, these works do not take data into consideration, \cite{krahenbuhl2015data} first introduce a data-driven handle structured initialization. Furthermore, \cite{mishkin2015all} proposed layer sequential data-driven initialization for the very deep neural network with faster convergence

Another line of related work is Information Bottleneck Theory and its application. The Information Bottleneck mainly studies how neural network generalizes and its training dynamic by estimating the mutual information of latent representation with $X, Y$. DVIB \cite{alemi2016deep} first applies IB of practice use as a regularizer which shows more robust representation. IB is then widely used for improving the generalization ability in various domains, such as GNN \cite{wu2020graph}, GAN \cite{peng2018variational, belghazi2018mutual}, Image Recognition \cite{hjelm2018learning} and so on.

%% file: model.tex
\section{Models}
In this section, we describe \short{} in detail. As our main idea is to select winning neurons from a large set of candidates, the key is to design the candidate neuron evaluation criterion based on IB. We summarize the insights of IB from two perspectives, which are encoded as our mutual information based initialization principle detailed in Sec. \ref{sec:principle}.  To make it efficiently calculable, we introduce detailed criteria to simplify mutual information calculation in section  \ref{sec:simplification}. We introduce the neuron campaign algorithm to perform the initialization in Sec. \ref{sec:Campaign}.

\subsection{IB based Initialization Principle}
\label{sec:principle}
In this section, we introduce two principles induced from different perspectives of IB. From the perspective of training dynamics, there exists a rapid increase in both input information maintenance and target-related information enhancement in the early training phase according to the IB theory.  
Therefore, we recognize the first principle that initialization should maximize the mutual information of $\textbf{Z}$ with both $\textbf{X}$ and $\textbf{Y}$ because initialization can be viewed as the early phase of training.  
Mathematically, to maximize mutual information of $\textbf{Z}$ with $\textbf{X}$:
\begin{equation}
    \label{equ:X_opt_basic_criterion}
        \begin{split}
        & \max_{\mathbf{W}} \  I\left ( \mathbf{X};\mathbf{Z}\right )\\
         \mathrm{s.t. }  &   \quad \forall j,\;  \left \| \mathbf{W}_{\cdot,j}^{(i)} \right \|_2 < \varepsilon ,\quad \mathbf{W}=\{\mathbf{W^{(i)}}|i=1,2,\ldots,D\} 
    \end{split}
\end{equation}
where $\mathbf{W_{\cdot,j}^{(i)}}$ indicates the weight of $j_{th}$ neuron of $i_{th}$ layer, $I\left (\mathbf{X};\mathbf{Z}\right )$ represents the mutual information between $\mathbf{X}$ and $\mathbf{Z}$. $\mathbf{Z}$ is the output features through corresponding weight $W$. The norm of $W$ is bounded by a small constant $\varepsilon$ to avoid the undesired increase in the mutual information due to the effect of the large norm.
Similarly, we intent to maximize the mutual information of $\textbf{Z}$ with $\mathbf{Y}$:
\begin{equation}
\label{equ:Y_opt_basic_criterion}
    \begin{split}
    & \max_{W} \  I\left ( \textbf{Z};\textbf{Y} \right ) \\
   \mathrm{s.t. }  &  \quad \forall j,\;\left \| \mathbf{W}_{\cdot,j}^{(i)} \right \|_2 < \varepsilon ,\quad \mathbf{W}=\{\mathbf{W^{(i)}}|i=1,2,\ldots,D\}
    \end{split}
\end{equation}
From the perspective of the network architecture, the second principle from IB suggests that $I(\mathbf{X};\mathbf{Z})$ plays an important role in the front layers while $I(\mathbf{Z};\mathbf{Y})$ is emphasized more on the rear layer close to the model output.
Hence, we enlarge mutual information between $\mathbf{X}$ and $\mathbf{Z}$ at the front layers and pay more attention to the mutual information between $\mathbf{Z}$ and $\mathbf{Y}$ on the rear layer.

To simultaneously optimize both targets on latent representations following the aforementioned principle, we combine  Eq.\eqref{equ:X_opt_basic_criterion} and Eq. \eqref{equ:Y_opt_basic_criterion} at each layer of a neural network as: 
\begin{equation}
    \label{equ:overall_perform}
    \begin{split}
    & \max_{W}\sum_{d=1}^D \alpha_{i} I\left ( \mathbf{X};\mathbf{Z}_d \right ) + \left ( 1 - \alpha_{i}  \right ) I\left ( \mathbf{Z}_d; \mathbf{Y} \right ) \\
     \mathrm{s.t. }  &  \quad \forall j,\; \left \| \mathbf{W}_{\cdot,j}^{(i)} \right \|_2 < \varepsilon ,\quad \mathbf{W}=\{\mathbf{W^{(i)}}|i=1,2,\ldots,D\} 
    \end{split}
\end{equation}
where $D$ is the number of layers and $\alpha_{i}$ is the control parameter of $d$-th layer for balancing two kinds of mutual information.
The parameter is set larger on the front layers to pay more attention to reconstruct $\mathbf{X}$, while the parameter is smaller (so  $1-\alpha_{i}$ is bigger) on the rear layer to emphasize more on the relevant information with $\mathbf{Y}$ by large $1-\alpha_{i}$.
However, it is hard to directly optimize Eq. \eqref{equ:overall_perform} for: (1) it is targeted at parameters of all layers; (2) it is difficult to efficiently obtain exact computation of mutual information.  

\subsection{Mutual Information Simplification}
\label{sec:simplification}
In this section, we simplify the Eq. \eqref{equ:overall_perform} from two perspectives: 
(1) transform the initialization target of the entire neural network into a layer-sequential initialization; 
(2) estimate the mutual information $I(\mathbf{X};\mathbf{Z})$ and $I(\mathbf{Z};\mathbf{Y})$ with proper simplification.

A $D$-layers DNN can be regarded as a Markov chain as $\mathbf{X}\to \mathbf{Z}_1\to \mathbf{Z}_2\ldots \to \mathbf{Y}$.
When we estimate the mutual information on such a Markov chain, we can find that (1) the mutual information on $\mathbf{Z}_{i}$ relies on previous latent representations $\mathbf{Z}_{i-1}$: $\mathbf{Z}_{i}=f(\mathbf{Z}_{i-1})$;
(2) the mutual information $I(\mathbf{X};\mathbf{Z}_{i})$ satisfies the following Data Processing Inequality chain:
\begin{equation}
    H(\mathbf{X}) \ge I(\mathbf{X};\mathbf{Z}_1) \ge I(\mathbf{X};\mathbf{Z}_2) \ge \cdots \ge I(\mathbf{X};\mathbf{Z}_D) \ge I(\mathbf{X};\mathbf{\hat{Y}}).  \label{equ:MC}
\end{equation}
The above observations indicate that the mutual information of the subsequent layers is bounded by the one of the current layer. In other words, if we want to maximize $I(\mathbf{X};\mathbf{Z}_{i})$, we should first ensure that $I(\mathbf{X};\mathbf{Z}_{i-1})$ is large enough. In addition, $\mathbf{Z}_{i}$ is also dependent on $\mathbf{Z}_{i-1}$. Therefore, it is reasonable to transform Eq.\eqref{equ:overall_perform} into a layer-wise optimization manner to sequentially generate latent representations for each layer: 
\begin{equation}
    \begin{split}
    \max_{\mathbf{W}_{i}}\  \alpha_{i}& I\left ( \mathbf{X};\mathbf{Z}_{i} \right ) + \left ( 1 - \alpha_{i}  \right ) I\left ( \mathbf{Z}_{i}; \mathbf{Y} \right ) \\
       \mathrm{s.t. }   \quad \forall j,\; \left \| \mathbf{W}_{\cdot,j}^{(i)} \right \|_2 < \varepsilon & ,\quad \mathbf{W}=\{\mathbf{W^{(i)}}|i=1,2,\ldots,D\}  
    \end{split}
\end{equation}
As we know that the mutual information is a key quantity across data science, it is difficult to estimate especially for continuous data.
We introduce simplification to estimate mutual information efficiently.
From the definition of mutual information, $I(X;Z_{i})$ can be expressed as:
\begin{equation} 
    \label{equ:primary_X}
 I(\mathbf{X};\mathbf{Z}_{i}) =  H(\mathbf{Z}_{i}) - H(\mathbf{Z}_{i}|\mathbf{X})  
 \end{equation}
where $H(\mathbf{Z}_{i})$ stands for the entropy of the latent representation $\mathbf{Z}_{i}$ and $H(\mathbf{Z}_{i}|\mathbf{X})$ denotes the conditional entropy.
Since the neural network is deterministic and there are finite instances in the dataset, the conditional entropy equals 0 and  Eq. \eqref{equ:primary_X} can be rewritten as:
\begin{equation}
    \label{equ:second_X}
 I(\mathbf{X};\mathbf{Z}_{i}) =  H(\mathbf{Z}_{i}) 
 \end{equation}
It is evident that on a given distribution like Gaussian distribution, larger standard deviation leads to larger value of entropy. Therefore, $H(\mathbf{Z}_{i})$ can be roughly estimated by:
\begin{equation}
\label{eq:criterion_1}
\begin{split}
     tr(\mathbf{\Sigma}_{i} )
\end{split}
\end{equation}
where $\mathbf{\Sigma}_{i}$ is the covariance matrix of $\mathbf{Z}_{i}$. 
Eq. \eqref{eq:criterion_1} could be considered as the \textbf{input information maintenance criterion}. 
Intuitively, it is similar to the Principal Component Analysis (PCA) which maintains information along the direction with the largest variance for avoiding large information loss on dimensionality reduction. 

$I(\mathbf{Z}_{i};\mathbf{Y})$ is highly correlated with the accuracy, i.e., the discrimination performance. Accordingly, to measure the discrimination, we simplify the estimation of $I(\mathbf{Z}_{i};\mathbf{Y})$ by calculating the intra-class variance and inter-class variance.
We introduce the mathematical form denoted as \textbf{target-related information enhancement criterion}:
\begin{equation}
tr \left ( \mathbf{\hat{H}}\mathbf{\hat{H}}^T \right ) -\frac{1}{N} \sum_{j=0}^N tr \left (  \mathbf{\hat{Z}}^T_{i} \mathbf{\Pi}^j \mathbf{\hat{Z}}_{i}  \right )
\end{equation}
where $\mathbf{\Pi} = \{ \mathbf{\Pi}^j \in \mathbb{R}^{m \times m} \}^N_{j=1}$ is a set of diagonal matrices whose diagonal entries encode the membership of the $m$ samples in $N$ classes. The diagonal entry $\mathbf{\Pi}^j_{i,i}=1$ indicates the $i_{th}$ sample belongs to the $j_{th}$ class. 
$H$ is the mean of representation on each class where $\mathbf{H}_{i,j} = \frac{1}{m_j} \sum_{k=0}^m{\left( \mathbf{Z}\mathbf{\Pi} \right )_{i,k}}$, $m$ and $m_j$ represent the number of all sample instances and instances for class $j$, respectively. $\mathbf{\hat{H}}$ and $\mathbf{\hat{Z}}_{i}$ are normalized $\mathbf{H}$ and $\mathbf{Z}$ with zero mean value.
Mathematically, the discrimination criterion is positively correlated with $I(\mathbf{Z}_{i};\mathbf{Y})$ under the condition that the inputs belonging to class $i$ obey the Gaussian distribution, i.e., $\mathbf{X}_i\sim N\left ( \mathbf{\mu} _i, \mathbf{\Sigma}_i \right )$.
Intuitively, the discrimination criterion encourages both strong intra-class concentration of representation $\textbf{Z}$ and large distance between different classes. This is also closely related to Linear discriminant analysis (LDA) which attempts to force the data from the same class concentrated and the data from different classes dispersed as much as possible.

\subsection{Neuron Campaign}
\label{sec:Campaign}
In this section, we detail the new initialization algorithm called \textbf{Neuron Campaign initialization (NCI)} that adopts the aforementioned properties without the time-consuming continuous optimization phase. The key idea of NCI is to find the optimal neuron subset from a large set of randomly initialized neurons and then integrate them as the initialization weight.

The details of NCI are described as follows. 
We firstly initialize a large candidate neuron set $\mathbf{W}_{c} \in \mathbb{R}^{d_{i-1} \times kd_i}$ (each column represents a neuron) containing $k$ times the number of neurons ($d_i$) in the $i$-th layer. Then, to obtain latent representations of all candidate neurons $\hat{\mathbf{Z}}_{i} \in \mathbb{R}^{m \times kd_i}$, we feed the output of previous layer, i.e. $\mathbf{Z}_{i-1}$ as the input. $m$ is the number of samples. 
To obtain appropriate neuron weights, we design a campaign mechanism: for each neuron, we calculate the score $s \in \mathbb{R}^{kd_i}$ based on the input information maintenance and target-related information enhancement criteria illustrated in Eq. \eqref{equ:second_X} and Eq. \eqref{eq:criterion_1} respectively.
To constitute the optimal subset of neurons $\mathbf{W}'$, we iteratively select the neuron with the largest score, and the selection at the $t$-th iteration from the left  candidate neuron set $\mathbf{W}^t=\mathbf{W}_c\setminus \mathbf{W}'$ can be formulated as:
\begin{equation}
\label{equ:best_neuron}
    \begin{split}
        & \mathbf{w}^{*}_t = \mathbf{W}^t_{\cdot, i} \\ 
       \mathrm{s.t. }  &  \quad i=\arg\max_i\ s_i\\
    \end{split}
\end{equation}
where $\mathbf{W}^t_{\cdot, i}$ is the $i_{th}$ column of $\mathbf{W}^t$, and $\mathbf{w}^{*}_t$ is the weight of the selected neuron at the current step. The initialization weight formed by the winning neurons is updated through:
 \begin{equation}
      \mathbf{W}^{'} := [\mathbf{W}^{'},\mathbf{w}^{*}_t].
 \end{equation}
However, selecting neurons with the highest scores may result in great similarity in their weights, thus leading to similar representations and degradation of model performance. For example, symmetry initialization leads to all neurons perform the same calculation, which makes the whole network useless.
Therefore, we take into consideration the diversity of the selected neuron subset and expect the newly selected neuron to be more orthogonal. More precisely, the norm of remaining neurons should be large in the null space of the selected neurons. We detail the selection at Algorithm. \ref{algorithm}.
\begin{equation}
\label{eq:best_neuron}
    \begin{split}
        & \mathbf{w}^{*}_t = \mathbf{W}^t_{\cdot, i} \\ 
       \mathrm{s.t. } \quad & i=\arg\max_i\ s_i \cdot \frac{||\mathbf{p}_t^i||}{||\mathbf{W}_{i}||}
    \end{split}
\end{equation}
where $d_t^i$ is the orthogonal component of the neuron on the null space of the selected neurons. The score is scaled by the ratio of the norm of orthogonal component and origin norm.

Furthermore, we anlayze the time complexity of \short{}, which is only $ O \left ( \left ( m \times d_{i-1} \right ) \times kd_{i}  + \left ( d_{i-1} \times d_{i-1} \right ) \times kd_{i} \right)$. Roughly speaking, it is similar to that of $k$ forward step. $k$ is usually set as 3 or 5 empirically. Although its computational cost is low, we could further greatly reduce it by sampling a subset of training samples to perform our algorithm.  

\begin{algorithm}
\renewcommand{\algorithmicrequire}{\textbf{Input:}}
	\renewcommand{\algorithmicensure}{\textbf{Output:}}
    \caption{Neuron Campaign initialization algorithm}
    \label{algorithm}
    \begin{algorithmic}[1]
        \REQUIRE Candidate weight matrix $\mathbf{W}$
        \FOR{t=1 to T}
        \STATE Update generalized orthnormalization matrix at $t$ steps: $\mathbf{A}_t = (\mathbf{A}_{t-1}, \mathbf{a}_t^T)^T$
        \STATE Calculate the null space projection by $\mathbf{P}_t=\mathbf{P}_{t-1}-\mathbf{a}_t\mathbf{a}_t^T \mathbf{W}$
	\STATE Select optimal neuron whose index is chosen by $i=\max_i s_i \frac{||\mathbf{p}^i_t||}{||\mathbf{W}_{i}||}$ 
        \STATE Update $\mathbf{w}^*=\mathbf{W}_{\cdot, i}$
        \STATE Normalize basis of the generalized orthnormalization matrix as $\mathbf{a_{t+1}}=\mathbf{p}^{i}_t/|| \mathbf{p}^{i}_t||$
        \ENDFOR
        \ENSURE Winning neurons formed weight matrix $\mathbf{W}'$ 
    \end{algorithmic}
\end{algorithm}

%% file: experiment.tex
\section{Experiments}
\subsection{Experiment Setup}
\textbf{Baselines: } To evaluate the performance of the proposed initialization strategy, we compare it with baseline initialization methods, including two basic methods Xavier initialization \cite{glorot2010understanding} and He-initialization \cite{he2015delving} and LSUV \cite{mishkin2015all}.  
The initialization strategies are applied on three different MLPs with ReLU as the activation function: MLP-2 (784, 100, 10), MLP-3 (784, 256, 100, 10), MLP-5(784, 32, 32, 32, 32, 10). MLP-L denotes MLP with L layers.  MLP-2 represents the shallow network with limited expression. MLP-3 represents the frequently used network which is both expressive and easy to optimize. MLP-5 represents the thin and deep network which is hard to optimize.
The architectures of different networks are [100], [256, 100], [32, 32, 32, 32] respectively.

\textbf{Hyperparameter Setup: } We use SGD \cite{ruder2016overview} with a mini-batch of 100 and a learning rate of 0.1 for all experiments. The training epoch is set to 200 for convergence. We apply random search strategy to find the proper $\alpha$ from 0 to 1. Each result is average over 5 runs with different random seeds. 
\begin{table}[!ht]
\caption{minimal error rate and corresponding epoch comparison of \short{} with baseline methods on MNIST.}
\label{tab:basic}
\scalebox{0.8}{
    \begin{tabular}{l|c|ccc}
    \toprule
    Strategy & Layers & Vanilla & LSUV & \textbf{\short{}}\\
    \midrule
    \multirow{3}{*}{Xavier} 
    & 2 & 2.04 $\pm$ 0.03 (75) &  2.05 $\pm$ 0.06 (51) & \textbf{1.93 $\pm$ 0.06 (60)} \\
    & 3 & 1.82 $\pm$ 0.05 (52) &  1.80 $\pm$ 0.07 (63) & \textbf{1.71 $\pm$ 0.09 (36)} \\
    & 5 & 2.83 $\pm$ 0.16 (98) &  3.13 $\pm$ 0.17 (69) & \textbf{2.53 $\pm$ 0.09 (78)} \\
    \midrule
    \multirow{3}{*}{He} 
    & 2 & 2.03 $\pm$ 0.03 (65) &  2.00 $\pm$ 0.04 (70) & \textbf{1.93 $\pm$ 0.07 (57)} \\
    & 3 & 1.83 $\pm$ 0.05 (54) &  1.86 $\pm$ 0.07 (71) & \textbf{1.73 $\pm$ 0.04 (35)} \\
    & 5 & 2.76 $\pm$ 0.07 (80) &  2.90 $\pm$ 0.12 (77) & \textbf{2.62 $\pm$ 0.08 (73)} \\
    \bottomrule
    \end{tabular}
}
\end{table}
\subsection{Main Experimental Results}
We conduct experiments on MNIST \cite{lecun1998mnist} with not only the traditional initialization strategies, Xaiver and He initialization, but also \short{}, LSUV on top of two traditional strategies. We present experimental results of the minimal error rate and its corresponding epoch to verify the generalization ability and the convergence speed, respectively.
The performance of traditional methods is not competitive and He initialization does not outperform Xaiver initialization as expected. The reason for this observation is that these methods assume that both inputs and weights are i.i.d. ideally, which is inconsistent with the practice use.
Different architectures also show varied performances. MLP-3 performs better than MLP-2 for better expressiveness. MLP-5 shows the worst result among all structures which indicates the difficulty of optimizing the deep and thin network.
\short{} consistently improve the performance of MLP with different layers: the relative error is reduced by 5.4\% at least and 10.6\% at most. Convergence speed is also improved as only 35 epochs are needed for convergence in the most frequently used MLP-3 as shown in Table \ref{tab:basic}. This observation may result from introducing labels with IB guidance which is detailed in Sec. \ref{sec:ablation}.
\begin{table}[!ht]
\caption{minimal error rate and corresponding epoch comparison of \short{} with methods with only one criterion.}
\label{tab:ablation}
\scalebox{0.8}{
    \begin{tabular}{l|c|ccc}
    \toprule
    Strategy & Layers & \short{} & TIE & IIM \\
    \midrule
    \multirow{3}{*}{Xavier} 
    & 2 & \textbf{1.93 $\pm$ 0.06 (60)} & 2.04 $\pm$ 0.07 (58) & 2.07 $\pm$ 0.09 (84) \\
    & 3 & \textbf{1.71 $\pm$ 0.09 (36)} & 1.82 $\pm$ 0.03 (43) & 1.82 $\pm$ 0.05 (52) \\
    & 5 & \textbf{2.53 $\pm$ 0.09 (78)} &  2.68 $\pm$ 0.05 (82) & 2.57 $\pm$ 0.09 (84) \\
    \midrule
    \multirow{3}{*}{He} 
    & 2 & \textbf{1.93 $\pm$ 0.07 (57)} & 2.07 $\pm$ 0.06 (59) & 2.034 $\pm$ 0.09 (62) \\
    & 3 & \textbf{1.73 $\pm$ 0.04 (35)} & 1.83 $\pm$ 0.07 (42) & 1.856 $\pm$ 0.05 (55) \\
    & 5 & \textbf{2.62$\pm$ 0.08 (73)} & 2.89 $\pm$ 0.11 (74) & 2.67 $\pm$ 0.12 (86) \\
    \bottomrule
    \end{tabular}
}
\end{table}

\subsection{Ablation Study}
\label{sec:ablation}
In this section, we design two strategies with only one criterion to figure out the character of individual criterion and verify the effectiveness of combining two criteria. 
The numerical results are shown in Table \ref{tab:ablation}.
where \dis{} is the strategy to select neuron only relied on the \textbf{T}arget \textbf{I}nformation \textbf{E}nhancement, and \std{} is similar which only relies on the \textbf{I}nput \textbf{I}nformation \textbf{M}aintenance.
Considering the error rate, \short{} consistently shows better results than \dis{} and \std{} which demonstrates the effectiveness of combining two criteria adaptively guided by IB. 
Then we analyze the character of each individual criterion.
\std{} shows comparable results with \short{} on the deeper MLP-5 which is hard to optimize. A potential reason is that the network with significant compression leads to input information loss. In such case, input information maintenance is of great need.
\dis{} leads to similar fast convergence speed with \short{}. This observation proves that better initial discrimination ability with label guidance may lead to a closer position near the local minima.  
On the other hand, the reason for slow convergence of \std{} may lie in that redundant input features at initialization take a long time for training to compress features irrelevant with $Y$.

%% file: conclusion.tex
\section{Conclusion}
In this work, we explore the initialization guided with the Information Bottleneck Theory and propose \short{} with efficient neuron campaign algorithm.

As for the future work, one direction is to conduct more experiments with \short{} to broader neural network architectures. 
Another direction is to explore other properties for initialization since the neuron campaign algorithm can adapt more diverse criteria.
One more direction is to explore the hyper-parameter setting strategy, such as adaptively and automatically set $\alpha$ with IB guidance at each layer.